\documentclass[conference]{IEEEtran}
\IEEEoverridecommandlockouts
\usepackage[noadjust]{cite}

\usepackage{amsmath,amssymb,amsfonts}
\usepackage{mathtools}
\usepackage{algorithmic}
\usepackage{graphicx}
\usepackage{textcomp}
\usepackage{subfig}
\usepackage{multirow}
\usepackage[table,xcdraw]{xcolor}

\def\BibTeX{{\rm B\kern-.05em{\sc i\kern-.025em b}\kern-.08em
    T\kern-.1667em\lower.7ex\hbox{E}\kern-.125emX}}
\begin{document}

\title{ReIL: A Framework for Reinforced Intervention-based Imitation Learning
}

\author{Rom Parnichkun\thanks{Rom Parnichkun and Matthew N. Dailey are with the Department of Information and Communication Technologies, Asian Institute of Technology, Pathum Thani, Thailand (E-mail: rom.parnichkun@gmail.com, mdailey@ait.asia)}, Matthew N. Dailey, and Atsushi Yamashita\thanks{Atsushi Yamashita is with the Department of Precision Engineering, The University of Tokyo, Tokyo, Japan (E-mail: yamashita@robot.t.u-tokyo.ac.jp). The research described here was conducted while the first author was on exchange at The University of Tokyo.}}


\maketitle

\begin{abstract}
Compared to traditional imitation learning methods such as DAgger and DART, intervention-based imitation offers a more convenient and sample efficient data collection process to users. In this paper, we introduce Reinforced Intervention-based Learning (ReIL), a framework consisting of a general intervention-based learning algorithm and a multi-task imitation learning model aimed at enabling non-expert users to train agents in real environments with little supervision or fine tuning. ReIL achieves this with an algorithm that combines the advantages of imitation learning and reinforcement learning and a model capable of concurrently processing demonstrations, past experience, and current observations. Experimental results from real world mobile robot navigation challenges indicate that ReIL learns rapidly from sparse supervisor corrections without suffering deterioration in performance that is characteristic of supervised learning-based methods such as HG-Dagger and IWR. The results also demonstrate that in contrast to other intervention-based methods such as IARL and EGPO, ReIL can utilize an arbitrary reward function for training without any additional heuristics.
\end{abstract}


\section{Introduction} \label{introduction}
Imitation learning (IL) empowers non-experts to train robotic systems for arbitrary tasks. Improvements in IL would support more widespread adoption of robotic technologies in the home and workplace, but IL methods often require large amounts of supervised training data \cite{bc} and/or employ unintuitive data collection approaches \cite{dart, dagger}, creating a bottleneck in  the adoption of these technologies.

A more natural approach to training of robotic agents is through supervisor \textit{interventions}, in which the supervisor (human operator) only intervenes in the agent's activity to provide corrections when it is approaching or has reached an undesirable state. The intervention approach is natural, because over time, it incrementally reduces the need for the supervisor to take over as the agent improves its performance, and it also reflects the way humans learn motor tasks such as driving a car with a coach. Furthermore, as opposed to imitation learning with supervisor-generated data only, intervention-based learning produces vast amounts of agent-generated data that could support it in further optimizing the agent’s policy. 

Reinforcement learning (RL) methods are designed specifically to optimize agents based on such agent-generated data. Advances in RL have recently led to several breakthroughs in creating superhuman agents for games such as Atari, Go, and Chess \cite{muzero} as well as new successes in controlling complex physical characters within simulated environments \cite{deepmimic}. However, the main disadvantage of RL methods for real environments is that they require vast amounts of agent-generated trial-and-error data as they explore the environments. 

Another factor that may obstruct widespread adoption of robots is the task-specific feature engineering required in most robotic systems. This has been mitigated somewhat by the aforementioned advances in deep reinforcement learning, which have broadened the generality of neural network models for policies, value estimators, and planning, but because RL tasks often require a certain amount of past experience along with the current observation for agents to perform tasks effectively, different tasks may still require different model architectures, which may be difficult for non-experts to design.

In this paper, we propose a framework aimed at facilitating users in training real-world robot agents on multiple tasks with very little supervision and fine tuning using the following components.
\begin{itemize}
    \item A new intervention-based imitation learning (IIL) algorithm that combines IL (BC \cite{bc}) and RL (TD3 \cite{td3}) techniques to train agents on both supervisor-generated and agent-generated data concurrently.
    \item A general multi-task model capable of simultaneously processing demonstrations, supervisor corrections, past experience, and the current observation sequence.
\end{itemize}

In the rest of this paper, we provide an overview of related work, formulate the IIL problem, describe our solution, describe experiments to validate the approach, then conclude with perspectives on the future potential of ReIL.

\section{Related Work} \label{related-work}

We build upon and address some of the limitations present in other proposed IIL algorithms such as HG-DAgger \cite{hg_dagger}, ``Learning to Drive in a Day" (which we abbreviate L2D) \cite{l2d}, EIL \cite{eil}, EGPO \cite{egpo}, IWR \cite{iwr}, and IARL \cite{iarl}.

An IIL environment generates two types of data, namely supervisor-generated data and agent-generated data. Some IIL algorithms train agents with only supervisor-generated or only agent-generated data. For instance, L2D utilizes RL to optimize a lane keeping agent by rewarding total distance traveled before supervisor intervention. The authors were remarkably able to train a functional lane keeping agent in a real driving environment in just 30 minutes. However, in our experiments, we have found that adding IL on top of the L2D approach greatly improves the rate at which it learns a task. On the other hand, HG-DAgger has been shown to greatly improve the performance of agents in terms of the amount of supervisor corrections required, compared to standard behavior cloning (BC) \cite{bc} and DAgger. HG-DAgger works by simply imitating supervisor corrections with BC. However, because HG-DAgger relies solely on supervisor-generated data, the agent would not continue to learn when it is performing well enough to execute without supervisor corrections. Furthermore, any IIL method that only utilizes supervised learning, including HG-DAgger and IWR, may require agents to be pre-trained with BC to produce an initial state visitation distribution sufficient for further intervention-based training, as the agent would not learn to avoid the actions that lead to interventions directly.

Besides not using all available data for learning, another specific limitation of several current state of art methods is that they either do not use or misuse estimates of cumulative discounted future reward. For example, EGPO and IARL are intervention-based RL approaches that successfully optimize an agent for navigation tasks. However, unlike L2D, these methods lack episode termination upon supervisor intervention, breaking the critical assumption that the value of a particular state-action pair is the expected cumulative future reward that would be obtained by the agent on executing that action. This introduces the possibility of a value overestimation bias for actions that lead to supervisor corrections. To make this clear, consider the case where the same reward function is applied to both supervisor-generated data and agent-generated data. Under such conditions, the supervisor may initially obtain much higher rewards than the agent, thereby creating a scenario where agent actions that lead to supervisor corrections have overestimated values. This would generally incentivize agents to ``cheat" by purposely generating undesirable actions that induce corrections. IARL avoids this problem by introducing a heuristic of differentiating the reward $r_t$ applied to the supervisor and the agent: \(r_t^{mix} = r_t - g_t\beta\), where $\beta$ is a positive constant and \(g_t = 1\) during supervisor corrections and $0$ otherwise. This differentiation prevents training on a single value function for both types of data. On the other hand, EGPO does permit the use of the same reward function for both supervisor and agent actions, but this permission leads to a requirement for an additional constrained optimization to minimize supervisor interventions.

Although methods such as IARL and EGPO utilize cumulative discounted future reward in ways that require additional heuristics to avoid undesirable behavior, eliminating value estimators entirely, as some other methods do, is also undesirable, as it delegates responsibility for comparing the relative quality of actions to the user. EIL is an IIL method that outperforms BC, DAgger, and also HG-DAgger. However, these methods do not utilize expected cumulative future reward, instead introducing a manually determined hyperparameter that determines the quality of each data-point for the process of optimization.

\section{Problem Formulation} \label{problem-formulation}

In this section, we describe the IIL problem in terms of two objectives, a formal IIL objective and an informal system design objective.

\subsection{Intervention-based Imitation Learning Objective}

The agent $\pi_{\theta}$ is assumed to operate within an environment given as a tuple $(S,A,P,R^{task},\pi_s)$, where $S$ and $A$ are the sets of possible states and actions, $P$ and $R^{task}$ denote the transition probability function and the task-specific reward function, and $\pi_s$ denotes the policy of the supervisor. An IIL trajectory is a sequence of tuples $(s_t, a_t, f^{demo}_t) \in S\times A\times \{0,1\}$ for $t \in 1..T$. We assume that $a_t = (1-f^{demo}_t)\pi_{\theta^{t}}(s_t) + f^{demo}_t\pi_s(s_t)$, where $f^{demo}_t = 0$ for agent-generated data and $f^{demo}_t = 1$ for supervisor-generated data. Moreover, the supervisor is assumed to have a set of acceptable state-action pairs for a task $\mathcal{D}_{good} \subseteq S \times A$ and that when $(s_t, \pi_{\theta^{t}}(s_t)) \not\in \mathcal{D}_{good}$, the supervisor intervenes and corrects the agent.

In such situations, would like an agent capable of performing arbitrary tasks optimally in the sense of the following objective.
\begin{equation} \label{eq:objective}
\begin{gathered}
    \mathop{\text{max}}_{\theta}\Big(\mathop{\mathbb{E}}_{s_{k+1} \sim P(s_k, \pi_\theta(s_k))}[\sum_{k=0}^{T-1}{\gamma^{k}R^{task}(s_{k+1}, \pi_{\theta}(s_{k+1}))]\Big)},\\ 
    \text{subject to}\; s_k \in \mathcal{D}_{good},
\end{gathered}
\end{equation}
where $\gamma$ $[0, 1)$ is the discount factor. We also note the assumption of IIL that only $A$ is known to the agent; information about $S$, $\mathcal{D}_{good}$, $P$, $R^{task}$, and $\pi_{s}$ must be gathered through exploration.

Although an optimal agent would not require supervisor interventions at all, explicitly optimizing to prevent such interventions does restrict the agent's state visitation to $s \in \mathcal{D}_{good}$. This restriction, by limiting the scope of exploration for the agent, actually eases its optimization.

Hence, we define the IIL objective to be the maximization of the agent's task-specific performance simultaneous with preventing supervisor interventions. Furthermore, the supervisor corrections are assumed to bring the agent’s state to $s \in \mathcal{D}_{good}$ as soon as feasible. Therefore, we additionally add an objective to imitate the supervisor corrections where $s \not\in \mathcal{D}_{good}$, as this may further support the satisfaction of Equation (\ref{eq:objective}).

\subsection{System Design Objective}

Finding an agent satisfying Equation (\ref{eq:objective}) requires a training algorithm and a policy model. For IIL problem settings in the real world, it is important that both of these components facilitate users in training agents. Because training robots in real environments can be labor intensive and/or expensive, we focus on maximizing the sample efficiency of the framework by developing an algorithm that can be trained offline. Furthermore, so that different policy models are not needed for different tasks, we develop a multi–task model that can reduce or eliminate task-specific feature engineering.

\section{Approach} \label{approach}

\subsection{Intervention-based Imitation Learning Algorithm}

Given $(S,A,P,R^{task},\pi_s)$, we aim to define a corresponding Markov decision process (MDP) $(S,A,P,R)$ that can be solved using RL techniques. That is to say, we reduce IIL to RL by designing a reward function $R$ that, when optimized by a RL agent, will ultimately satisfy Equation (\ref{eq:objective}).

First, we tackle the aforementioned problem of possible value overestimation bias in IIL by defining a surrogate IIL objective function that is the expected cumulative discounted future reward up to the point of supervisor intervention or task termination as follows. 
\begin{equation} \label{eq:surrogate_objective}
\begin{gathered}
    Q(s_t,a_t) = \mathop{\mathbb{E}}_{s_{t+1} \sim P(s_t, \pi_\theta(s_t))}[\sum_{k=0}^{K-t}{\gamma^{k}R(s_{t+k},a_{t+k})].}
\end{gathered}
\end{equation}

We define $K = \text{min}(T, \text{min}(\{t^{int} \in t..T \mid f^{demo}_{t^{int}+1} = 1\}))$. Besides preventing value overestimation bias that may incentivize the agent to purposely ``cheat”, as in EGPO, this formulation also allows the value estimator to be trained on both agent and supervisor generated data, as the only possible value estimation bias would be underestimation of the values of supervisor actions. 

To find $\pi_\theta$, we will utilize the actor-critic RL formulation in which a critic with parameters $\phi$ estimates the value in Equation (\ref{eq:surrogate_objective}) utilizing the Bellman equation
\begin{equation}
         Q_{\phi}(s_t,a_t) = R(s_t,a_t) + \gamma\big(1-\Omega^{mix}_{t}\big)Q_{\phi}\big(s_{t+1}, \pi_{\theta}(s_{t+1})\big),
\end{equation}
where
\begin{equation}
\begin{gathered}
    \Omega^{mix}_t = \text{max}(\Omega^{int}_t, \Omega^{task}_t), \\
    \Omega^{int}_t = \text{max}(f^{demo}_{t+1}-f^{demo}_t, 0).
\end{gathered}
\end{equation}

$\Omega^{mix}_t$, $\Omega^{int}_t$, and $\Omega^{task}_t \in \{0,1\}$ are all binary flags raised to effectively terminate the episode by restricting the value at $t$ to only $R(s_t,a_t)$. 

Next, in order to encourage the agent to minimize supervisor interventions, we design reward function $R$ as
\begin{equation} \label{eq:reward_func}
    R(s_t,a_t) = (1-\Omega^{int}_t)R^{task}(s_t,a_t) + \Omega^{int}_tr^{int},
\end{equation}
where $r^{int}$ is a constant satisfying 
\begin{equation} \label{eq:reward_constraint}
    r^{int} < \mathop{\text{min}}_{s\in S,a \in A}(R^{task}(s,a))/(1-\gamma).
\end{equation}

$r^{int}$ is the reward given upon supervisor intervention, and $R^{task}$ is any arbitrary reward function for the given task. For example, $R^{task}$ could be the magnitude of the velocity given by $a_t$, as it is in L2D. Note that $r^{int}$ must be lower than the minimum value attainable from $R^{task}$; which we assume to be the value of an action leading to an infinite sequence of $\text{min}(R^{task})$ reward. This means that the proposed MDP will lead to an agent that minimizes supervisor interventions, as the value of any action leading to an intervention now becomes strictly lower than an action that does not. 

Furthermore, we claim that optimizing the proposed MDP is strictly equivalent to optimizing Equation (\ref{eq:objective}) for any sub-trajectory that does not lead to supervisor intervention. This means that as long as intervention is avoided, our proposed MDP is a standard MDP, making our approach compatible with maximization of performance against any $R^{task}$. In order to solve this MDP, we utilize TD3 \cite{td3}, a RL algorithm for deterministic policies in continuous action spaces.

However, as mentioned earlier, solely utilizing RL has several disadvantages in real environments, as it requires random exploration of the environment. We therefore introduce BC to augment the exploratory learning with imitation of supervisor corrections. We modify the TD3 objective function as follows.
\begin{multline}
    J(\theta) = \mathop{\sum}_{(s,a,f^{demo}) \sim \mathcal{D}_{mem}}\Big[\alpha \underbrace{Q_{\phi}\big(s,\pi_{\theta}(s)\big)}_\text{TD3 objective} -\\
    f^{demo}\underbrace{||\pi_{\theta}(s) - a||^2}_{\text{BC objective}}\Big],
\end{multline}
where $\mathcal{D}_{mem}$ is a set of state-action-controller tuples collected from training rollouts and $\alpha$ is a hyperparameter that can be tuned to balance the influence of the RL and BC objectives functions.

\begin{figure}
\centerline{\includegraphics[width=2.5in]{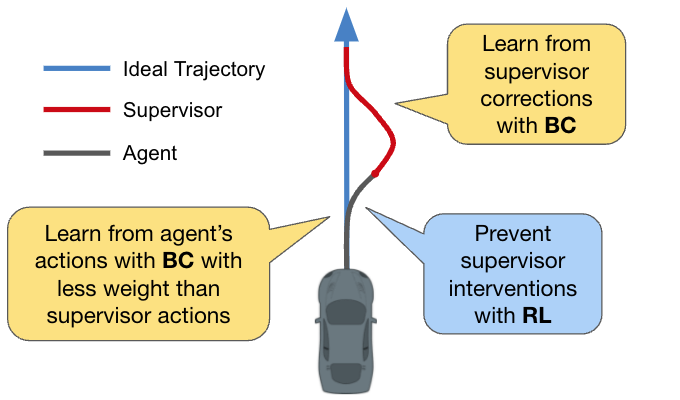}}
\caption{ReIL principles.}
\label{fig:ReIL-Overview}
\end{figure}

Previous work on offline RL that we abbreviate TD3+BC \cite{td3+bc} utilizes a similar objective function combining TD3 and BC allowing the off-policy RL algorithm to be trained offline. Offline RL is convenient in that it enables training of agents without the online feedback required for other RL algorithms. This is especially advantageous when training on real environments that make online training labor intensive and/or expensive. In TD3+BC, the BC term is applied at every timestep and helps regularize TD3, pushing the agent towards actions taken in the dataset without incentivizing unexplored actions whose values may be overestimated. However, naively utilizing the TD3+BC objective function for the task of IIL hinders the early trainability of the agent, as the agent initially produces poor actions that should not be imitated with BC. Therefore, we further modify the objective function by adding a hyperparameter $\beta < 1$  to allow BC for agent-generated actions with less weight than supervisor-generated ones as follows.
\begin{multline} \label{eq:obj-final}
    J(\theta) = \mathop{\sum}_{s,a,f^{demo} \sim \mathcal{D}_{mem}} \Big[\alpha Q_{\phi}\big(s,\pi_{\theta}(s)\big) - \\ 
    \big(f^{demo} +\beta(1-f^{demo})\big)||\pi_{\theta}(s) - a||^2\Big].
\end{multline}

Fig. \ref{fig:ReIL-Overview} illustrates the intuition behind Equation (\ref{eq:obj-final}).

\subsection{Multi-Task Imitation Learning Model}

The techniques developed in the previous section ensure that our IIL agents make effective use of both supervisor and agent generated actions. In this section, in order to ensure the best possible sample efficiency and generality of the framework, we develop a model capable of being conditioned on demonstrations, past experience, and current observations. This provides several theoretical advantages:
\begin{itemize}
    \item The model can be utilized for memory-dependent tasks, such as in obstacle avoidance, in which it is critical to remember the location of the obstacle currently being avoided, reducing task-specific feature engineering.
    \item The model has the capacity to take appropriate actions learned from past supervisor corrections.
    \item The same model can be trained on multiple tasks by conditioning with different user demonstrations, raising the possibility of generalizing and capturing similarities across different tasks.
\end{itemize}

In order to develop such a model, we adapt an existing meta-learning model to the task of imitation learning. Meta-learning is a field that focuses on creating agents that rapidly learn from new information. Effective meta-learning models are therefore capable of rapidly adapting to past experience to infer appropriate actions. SNAIL \cite{snail} is a meta-learning model that combines the advantages of both convolutional neural networks \cite{wavenet} and attention mechanisms \cite{transformer} by stacking these types of layers on top of one another. Moreover, SNAIL employs dense connections \cite{denseblock} across these layers, allowing the gradient to efficiently propagate throughout the entire stack of layers. These architectural choices make SNAIL suitable for temporal data such as the past experience of an agent. The SNAIL authors created an agent capable of efficiently navigating through complex mazes on second attempts by directly referencing  the past experience of states and actions executed in the first attempt. We note that demonstrations in IL and past experience in RL are qualitatively similar, making it sensible to adapt SNAIL to the problem of imitation. 

\begin{figure}
\centerline{\includegraphics[width=3.5in]{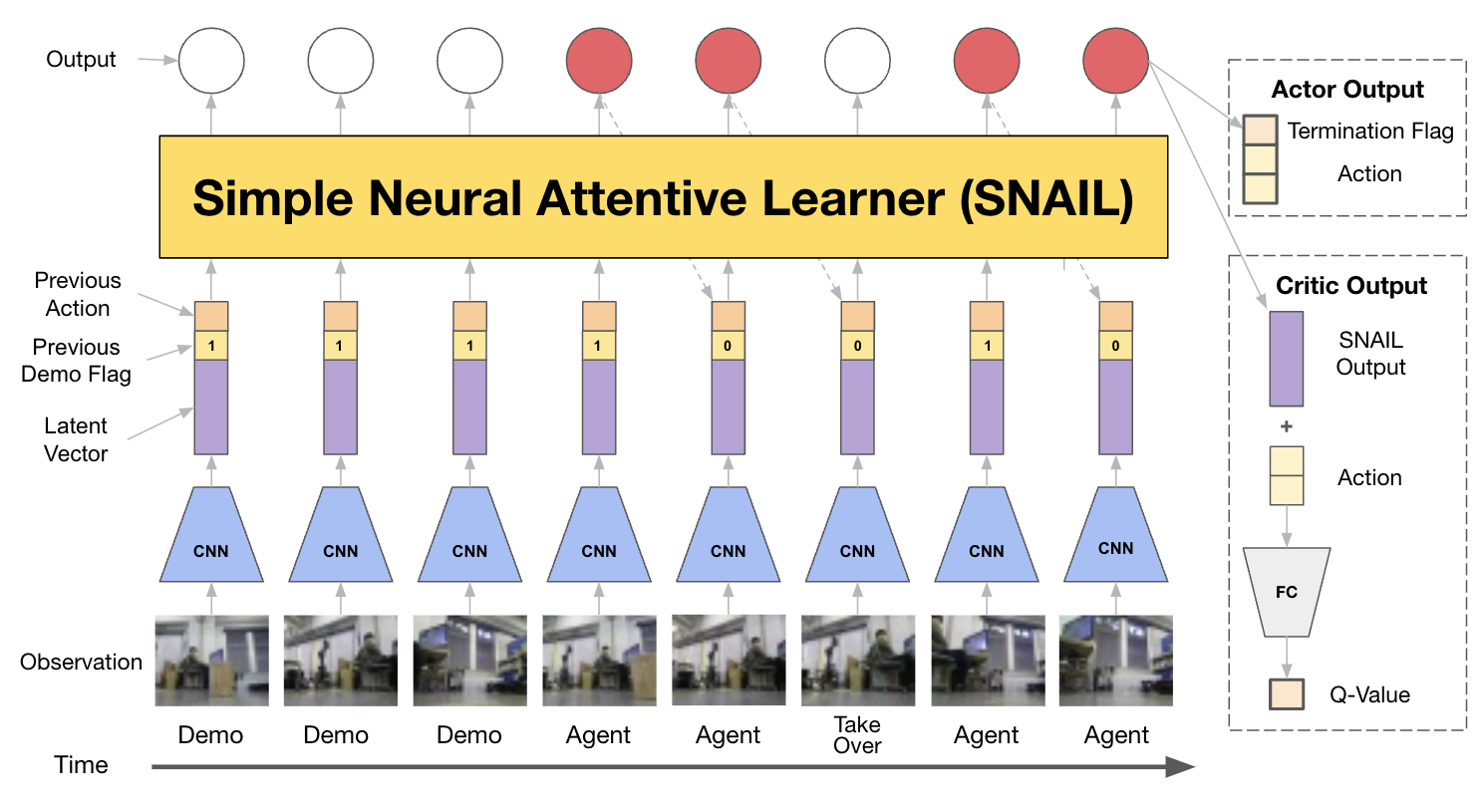}}
\caption{Illustration of the MimeticSNAIL model. The output is conditioned on the demonstration, agent actions, and supervisor corrections.}
\label{fig:MimeticSNAIL}
\end{figure}

Fig. \ref{fig:MimeticSNAIL} illustrates our proposed SNAIL IL model, which we call MimeticSNAIL. Two modifications were made to the original model architecture, as follows.
\begin{enumerate}
    \item Append the previous demonstration flag $f^{demo}_{t-1}$ to the latent vector fed to SNAIL.
    \item Augment the attention mechanism with a linear bias in the form of ALiBi \cite{alibi} as follows.
\end{enumerate}
\begin{equation}
    \text{Attention} = \text{softmax}\Big(\frac{\boldsymbol{QK}^T}{\sqrt{l_k}} - m|\boldsymbol{t}^T - \boldsymbol{t}| + \boldsymbol{M}\Big)\boldsymbol{V}.
\end{equation}

Here $\boldsymbol{Q}$, $\boldsymbol{V}$, $\boldsymbol{K}$, $\boldsymbol{M}$ are the query, value, key, and the causal mask matrices. The key matrix is $(T^{demo}+t) \times l_K$, $\boldsymbol{t}$ is a column vector of time frames, e.g. $[1,...,T^{demo},1,...,t]^T$, $m$ is a parameter included in $\phi$ or $\theta$, and $T^{demo}$ is the final timestep of the demonstration. This modification introduces a recency bias to the attention mechanism, and also increases the relative attention placed on demonstration time frames $t^{demo}$ in proximity to the queried time frame $\boldsymbol{t}$.

Additionally, the actor model’s output contains a task-termination flag denoted as $f^{tf}_\theta = \pi^{tf}_{\theta}(s)$, which, when raised, can be used to begin the next task automatically. This allow users to train the model on arbitrarily long tasks by queing a list of shorter sub-tasks. This sequencing can be learned by copying the supervisor's task termination signal $f^{s}$ provided during training and additionally including the following objective function in the overall objective for $\theta$.
\begin{multline}
    J^{tf}(\theta) = \mathop{\sum}_{(s,a,f^{demo}) \sim \mathcal{D}_{mem}}\big(1-f^{tf}_s\big)\log\big(1-\pi^{tf}_{\theta}(s)\big) + \\
    f^{tf}_s\log\big(\pi^{tf}_{\theta}(s)\big).
\end{multline}

\section{Experiments} \label{experiment}

We conducted experiments in simulation and real environments, aiming to compare the performance of different IIL algorithms with respect to supervisor burden. 

\subsection{Experimental Setup}

\begin{table}
\caption{Simulated Cartpole Algorithm Parameters}
\begin{center}
\resizebox{\columnwidth}{!}{
\begin{tabular}{ |c|c|c|c|c|c|c|c|c|c|c|c| } 
\hline
Algorithm & Batch & U/S & $\text{LR}_{\theta}$ & Decay & $\text{LR}_{\phi}$ & $\gamma$ & Delay & Noise & U/P & $\alpha$ & $\beta$ \\
\hline
ReIL, IARL & 24 & 50 & $10^{-6}$ & $10^{-4}$ & $10^{-4}$ & 0.99 & 0.005 & 0.2 & 2 & 0.05 & 0.1, 0.0\\
\hline
HG-DAgger & 24 & 50 & $10^{-6}$ & $10^{-4}$ & - & - & - & - & - & - & - \\
\hline
\end{tabular}}
\end{center}
\footnotesize{Batch = Mini-batch size, U/S = Updates per env. step, $\text{LR}_\theta$ = Actor learning rate, Decay = Actor weight decay, $\text{LR}_\phi$ = Critic learning rate, $\gamma$ = Discount factor, Delay = TD3 policy update delay parameter, Noise = TD3 target actor noise, U/P = TD3 actor update period}
\label{table:cartpole_algorithm}
\end{table}

\begin{table}
\caption{Mobile Robot Navigation Algorithm Parameters}
\begin{center}
\resizebox{\columnwidth}{!}{
\begin{tabular}{ |c|c|c|c|c|c|c|c|c|c|c|c| } 
\hline
Algorithm & Epoch & $\text{LR}_\theta$ & Decay & $\text{LR}_\phi$ & $\gamma$ & Delay & Noise & U/P & $\alpha$ & $\beta$\\
\hline
ReIL & 10 & $2(10^{-4})$ & $10^{-3}$ & $5(10^{-4})$ & 0.95 & 0.005 & 0.2 & 2 & 0.2 & 0.1\\
\hline
HG-DAgger & 10 & $2(10^{-4})$ & $10^{-3}$ & - & - & - & - & - & - & -\\
\hline
\end{tabular}}
\end{center}
\footnotesize{Epoch = Epochs per episode, $\text{LR}_\theta$ = Actor learning rate, Decay = Actor weight decay, $\text{LR}_\phi$ = Critic learning rate, $\gamma$ = Discount factor, Delay = TD3 policy update delay, Noise  = TD3 target actor noise, U/P = TD3 Actor update period}
\label{table:mobile_robot_algorithm}
\end{table}

We set up two environments: the OpenAI CartPole-V1 simulation and a real mobile robot equipped with a front facing monocular camera aimed at different indoor visual navigation tasks.

\subsubsection{Simulated Cart-pole}
The OpenAI CartPole-V1 environment was modified to support a continuous action space $A=[-1,1]$, and we utilized the original constant reward function $R^{task} = r^{int} = 1$, which satisfies Equation (\ref{eq:reward_constraint}). To simulate the supervisor in the simulation, we trained an agent to expert level with the standard TD3 algorithm in a non-IIL environment, and set the space of acceptable state-action pairs to $\mathcal{D}_{good} = \big\{(s, a) \in \mathbb{R}^4 \times A \mid |s_1|<2.0, |s_3| < 12\pi/180\big\}$. In the cartpole simulation, $s$ contains the cartpole's linear position, linear velocity, angle, and angular velocity, respectively. For this simple environment, we utilized a fully connected neural network for both the actor and the critic. We then compared ReIL, IARL,\footnote{Instead of the PPO algorithm \cite{ppo} used in IARL, we used TD3 to isolate the differences between our implementation of IARL and ReIL to only the reward function and objective function.} HG-DAgger, and ReIL with RL (TD3) only, which we denote as ReIL (Only RL). Note that ReIL (Only RL) is essentially the L2D algorithm with the cartpole reward function. The reward function for IARL is $R(s_t,a_t) = 1 - f^{demo}_t$. Additional algorithm-specific details are listed in TABLE \ref{table:cartpole_algorithm}.

The goal of the agent was to balance the cartpole for 3000 consecutive timesteps. As metrics, we measured the success rate and the total number of supervised steps required for the successful runs of each algorithm across multiple runs. 

\begin{figure}
\centerline{\includegraphics[width=1.6in]{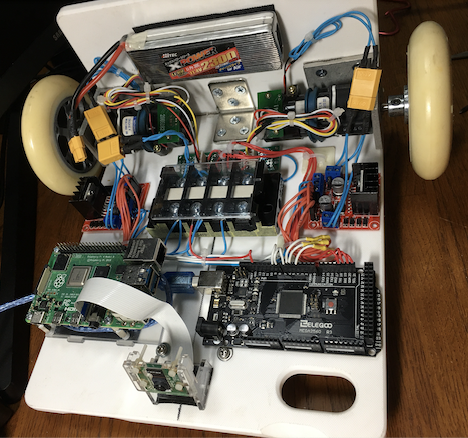}}
\caption{Indoor mobile robot platform for testing IIL methods.}
\label{fig:mobile_robot}
\end{figure}

\begin{figure}
    \centering
    \subfloat[\centering Obstacle avoidance task]{{\includegraphics[width=1.5in]{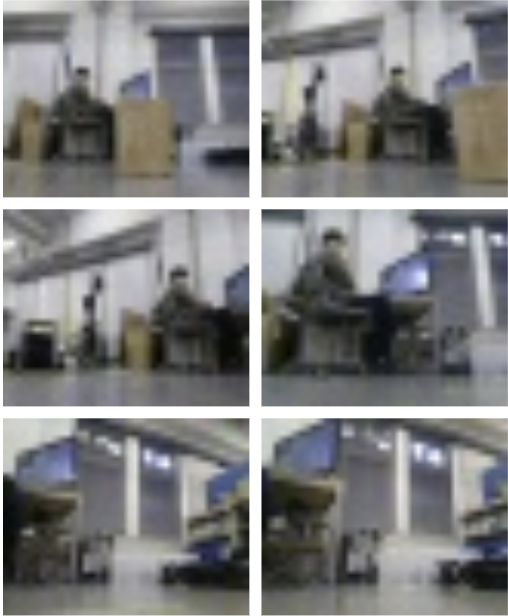} }}
    \qquad
    \subfloat[\centering Target hitting task]{{\includegraphics[width=1.5in]{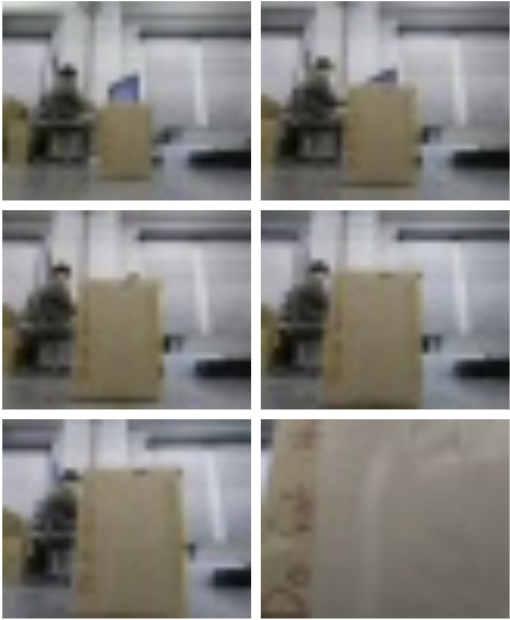} }}
    \caption{Sample sequences of agent observations.}
    \label{fig:observations}
\end{figure}

\subsubsection{Real World Mobile Robot Navigation}

To test and compare our algorithm in the real world, we developed a ROS2-enabled mobile robot equipped with a front facing monocular camera as shown in Fig. \ref{fig:mobile_robot}. The agent was set up as a high-level controller that outputs a vector of desired linear and angular velocities $a = \{ (v, \omega) \in \mathbb{R}^2 \mid |v| \leq 0.1 \; \text{m/s}, |\omega| \leq 0.4\;\text{rad/s}\}$ at 2Hz based on its visual input. The low-level controller converts the desired body velocities to the necessary wheel velocities and controls the wheels using an independent PI controller for each wheel and an outer-loop P controller for the wheel differential, all running at 100Hz. We lowered the image resolution to $40\times30$ RGB pixels to keep computational requirements modest. We formulated a universal reward function usable for any episodic IIL task as follows.
\begin{equation}
\begin{gathered}
R^{task}(s_t, a_t) = 1, t \in [1..T-1], \\
R^{task}(s_T, a_T) = 2/(1-\gamma), r^{int} = 0.
\end{gathered}
\end{equation}

We tested this reward function with tasks including obstacle avoidance and target hitting, with sample observations as shown in Fig. \ref{fig:observations}. We train the agents both online and offline; the agents trained offline utilized all of the data collected during the online training experiments. Each online training experiment was conducted in either one or two runs of 100 episodes.

For these difficult navigation tasks, we utilize two separate MimeticSNAIL models, one for the actor and one for the critic. Both models process the $40\times30$ pixel observations with a 16-channel then a 32-channel convolutional layer, both with $4\times4$ kernels and $\text{stride} = 2$, followed by two fully-connected layers with 100 units, outputting a 100-dimensional latent vector that is then fed into the SNAIL portion of the model. The SNAIL model uses a TCBlock(L, 30), an AttentionBlock(16, 16), and a TCBlock(L, 30) with L = 75.\footnote{Due to space limitations, we refer readers to SNAIL \cite{snail} for details on the TCBlock and the AttentionBlock.} The output of the SNAIL portion of the model is fed to an additional fully-connected layer to finally output either $(a, f^{tf})$ for the actor or $Q_\phi$ for the critic. The agent did not receive any demonstration before RL began, and on every episode, it faced a new obstacle or target configuration.

We mainly compare ReIL, ReIL (Only BC) and algorithms pre-trained with BC, including HG-DAgger and ReIL (Pre-trained). Further algorithm-specific details are listed in TABLE \ref{table:mobile_robot_algorithm}.

Because performance is more difficult to assess in noisy real world environments than in simulation, we measured the following four separate metrics.
\begin{enumerate}
    \item \textit{Average episode length}: The time it takes the agent to complete the task, which is a rough estimate of the agent's efficiency at completing the task.
    \item \textit{Average absolute angular acceleration}: An estimate of the smoothness of the agent-generated trajectory, which can highlight undesirable oscillatory behavior. Computed as $\frac{1}{T}\sum_{t=1}^T|\omega_{t}-\omega_{t-1}|$.
    \item \textit{Number of supervised steps}: An estimate of the supervisor's burden during online training or offline execution.
    \item \textit{Action error}: An estimate of the agent's performance in imitating the supervisor. The data are collected by having the supervisor provide DAgger-like off-policy desired actions during times that the agent is controlling the robot. Computed as the normalized RMSE between supervisor and agent actions ${\sqrt{\frac{1}{T}\sum_{t = 1}^{T}||\frac{\pi_\theta(s_t)-\pi_s(s_t)}{2a_{\text{max}}}||^2}}$ where $a_{\text{max}} = (0.1, 0.4)$.
\end{enumerate}

\begin{figure}
\centerline{\includegraphics[width=3.5in]{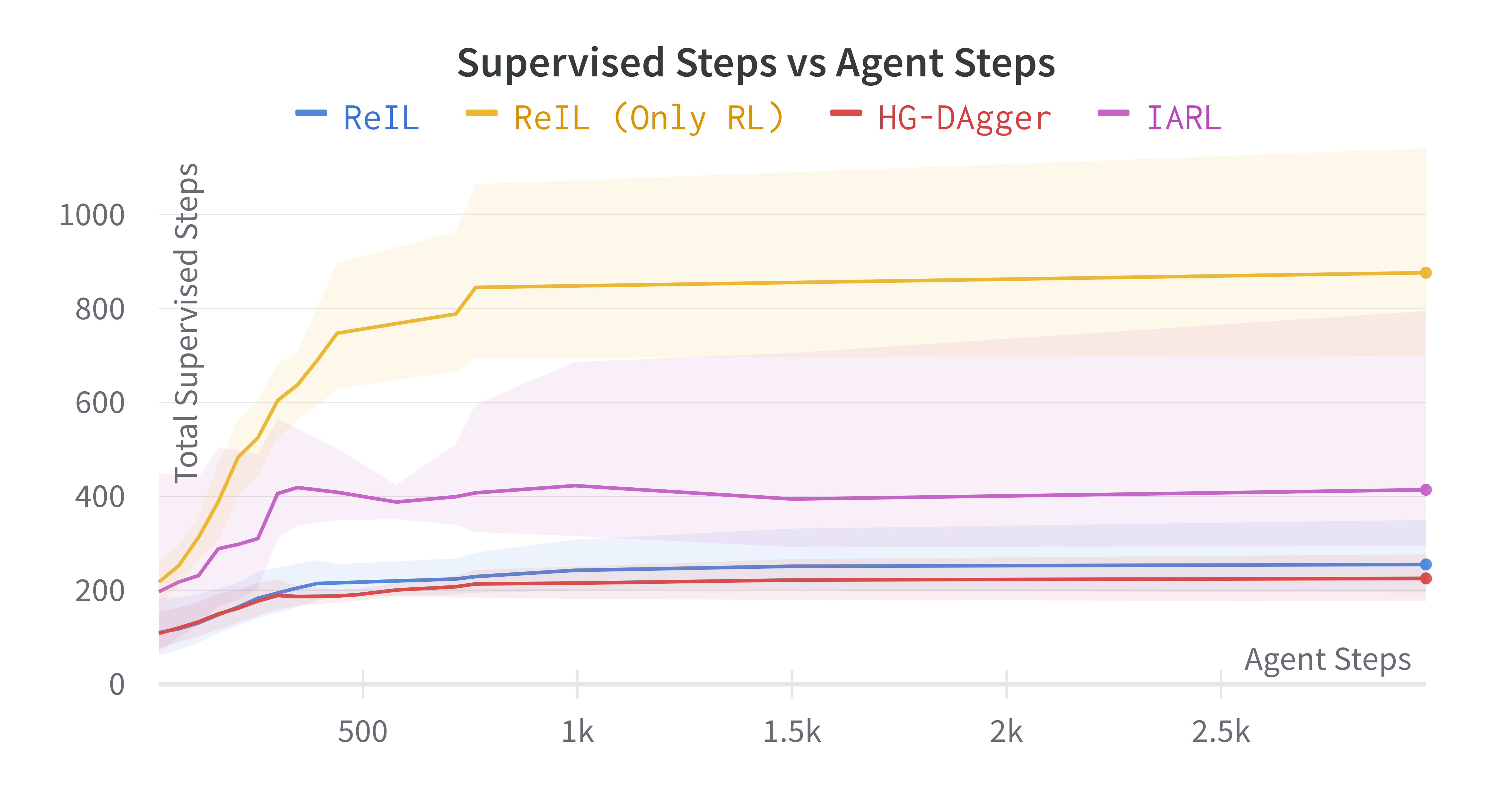}}
\caption{Simulation results. Total supervised steps required vs. the agent's average performance. A moving average filter of $n = 10$ has been applied.}
\label{fig:cartpole-result}
\end{figure}

\begin{figure}
    \centering
    \subfloat[\centering][The time required for the agent trained with ReIL (Only BC) increases over time.]{{\includegraphics[width=3.5in]{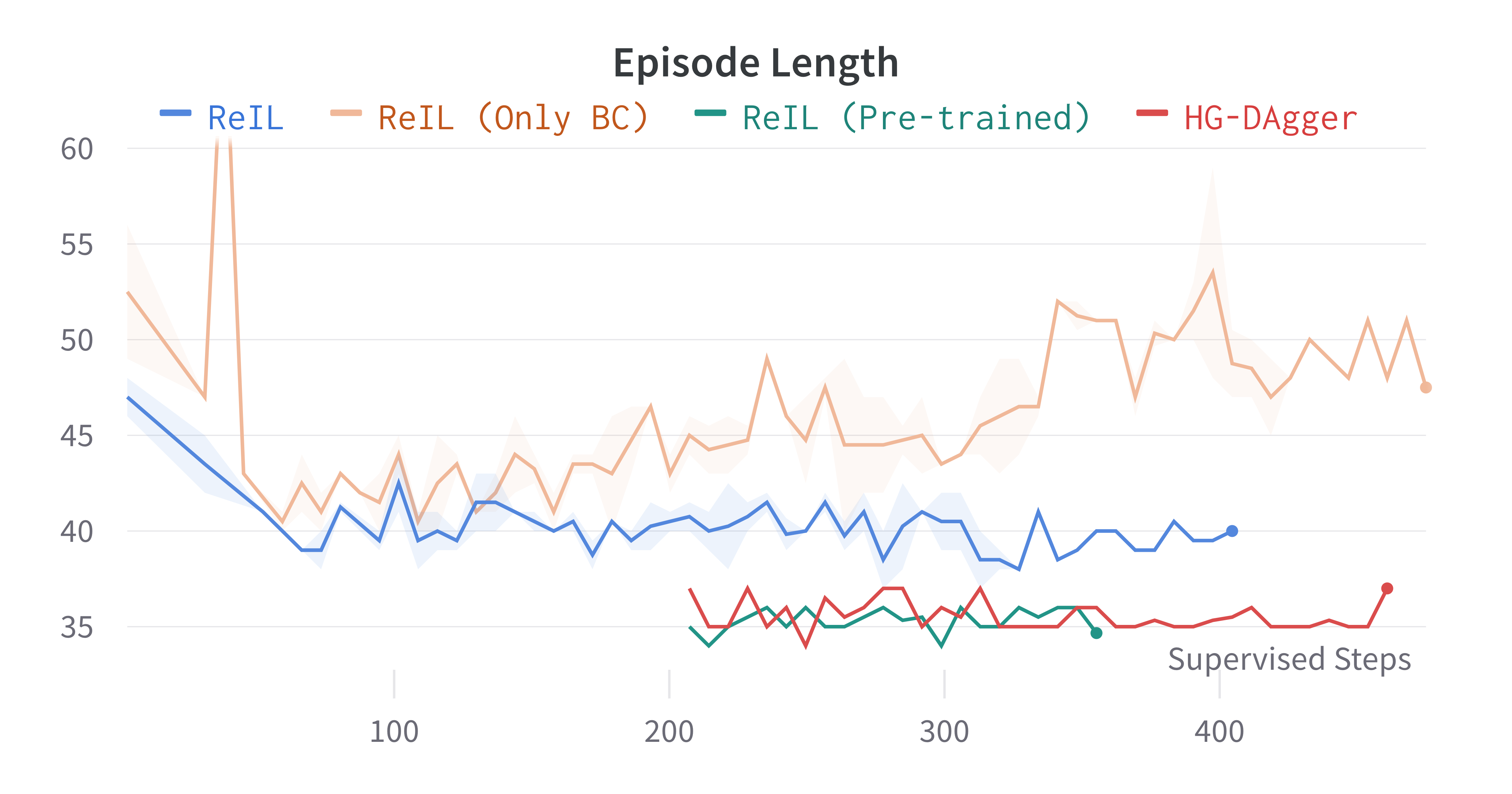} }}
    \qquad
    \subfloat[\centering][The HG-DAgger trained agent suffers from increased oscillation as supervisor corrections increase. ]{{\includegraphics[width=3.5in]{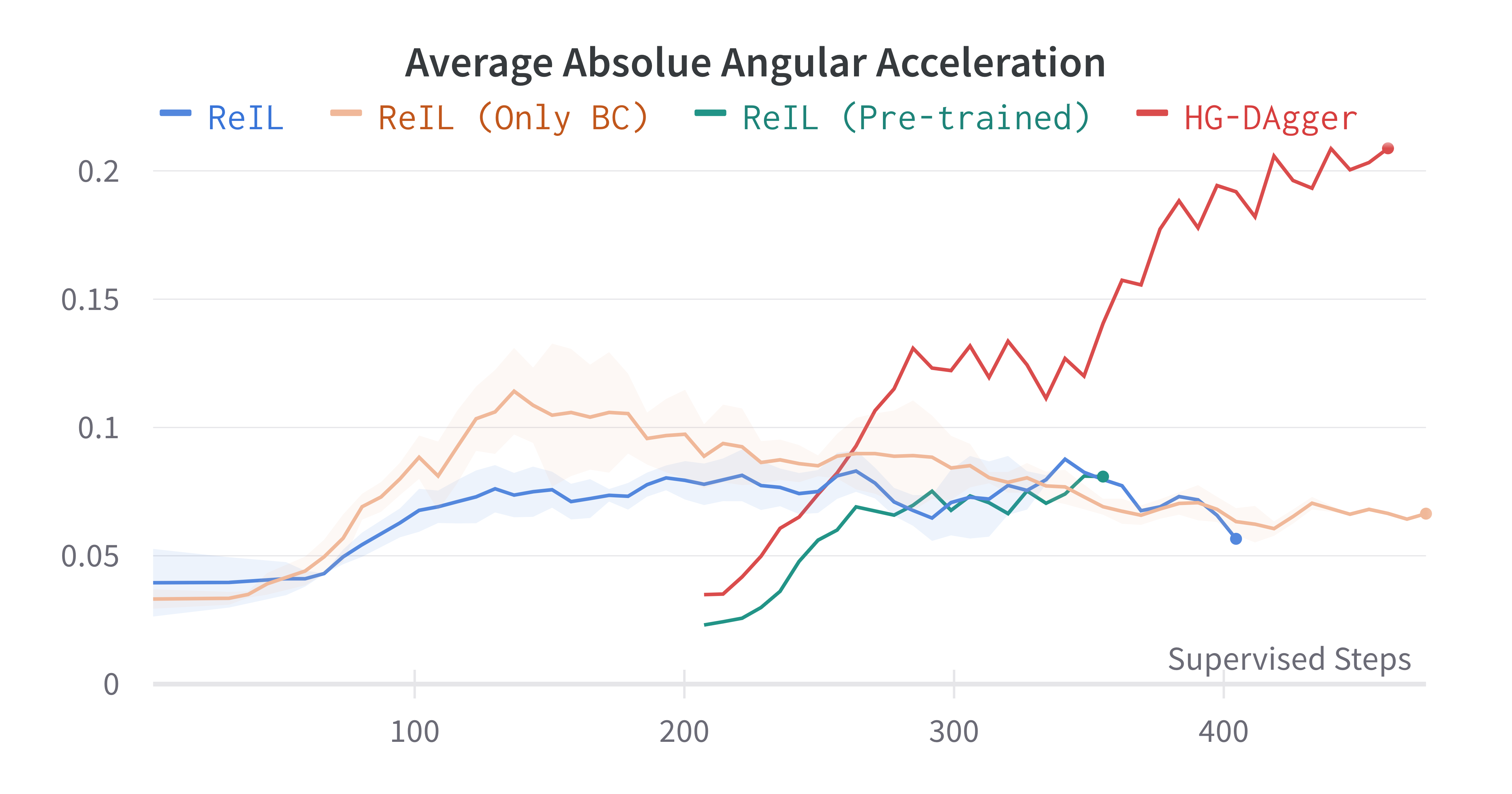}}}
    \qquad
    \subfloat[\centering][All algorithms lowered the average number of supervised steps per episode over time.]{{\includegraphics[width=3.5in]{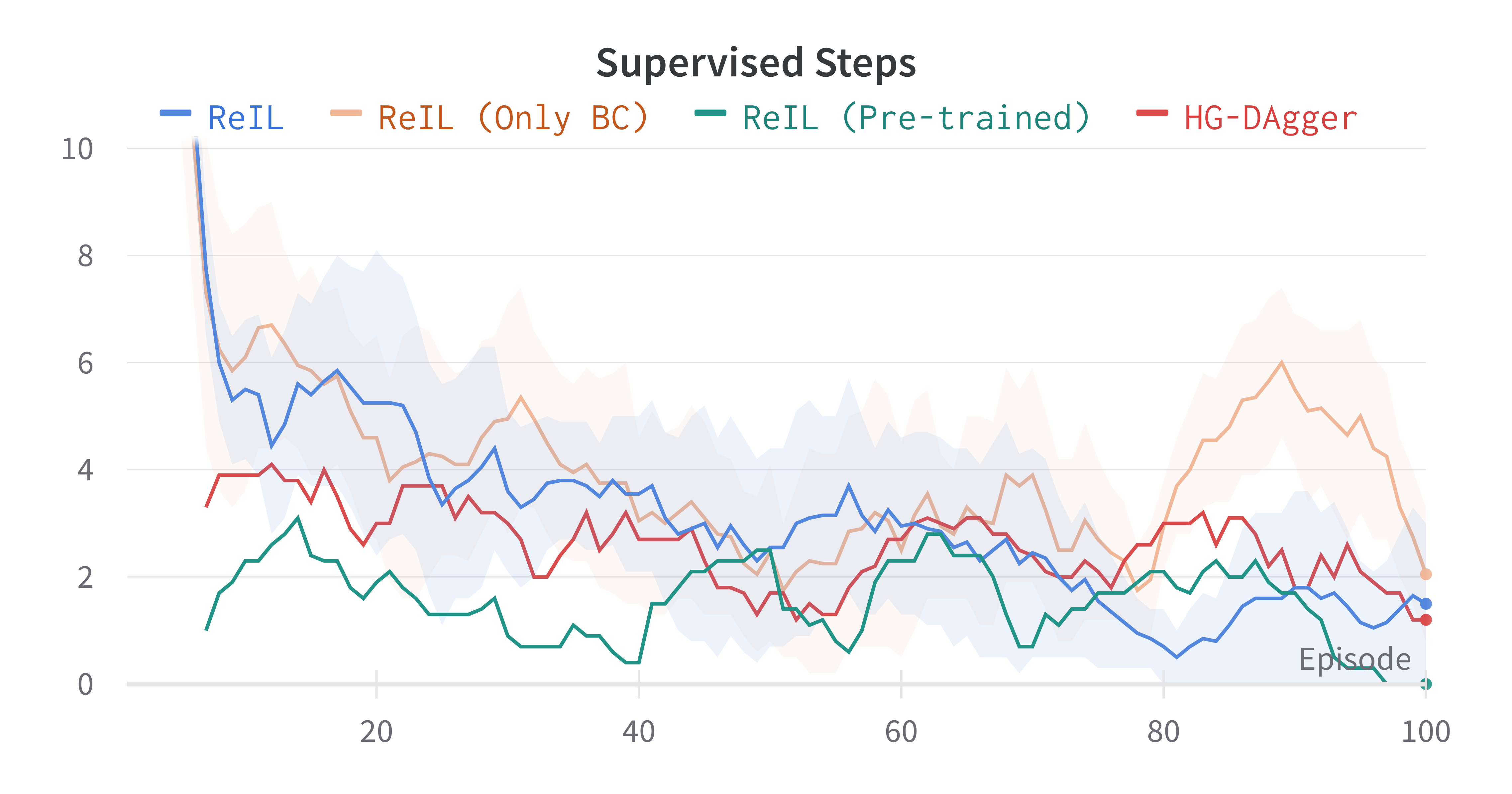} }}
    \caption{Obstacle avoidance results. A moving average filter of $n = 10$ has been applied to (b) and (c).}
    \label{fig:mobile-robot-result}
\end{figure}

We note here that comparisons with other algorithms reflect contrasts with specific details of these algorithms, not a complete implementation of the algorithm.

\begin{table}
\caption{Simulated Cartpole Results}
\centering
\begin{tabular}{ |c|c|c| } 
\hline
Algorithm & Supervised Steps & Success Rate\\
\hline
ReIL & \textbf{270.7 (53.21)} & $20/20$\\
\hline
ReIL (Only RL) & 978.667 (393.26) & $3/4$\\
\hline
HG-DAgger & \textbf{251.6 (45.144)} & $20/20$\\
\hline
IARL & 411.27 (267.731) & $16/20$\\
\hline
\end{tabular}
\label{table:cartpole_result}
\end{table}

\begin{table}[]
\caption{Mobile Robot Navigation Results}
\begin{center}
\resizebox{\columnwidth}{!}{
\begin{tabular}{|l|l|l|l|l|l|l|}
\hline
Task                   & Mode                        & Algorithm                  & Error                       & Supervised    & Steps                      & Angular Acc.                      \\ \hline
\multirow{6}{*}{OA} & \multirow{4}{*}{Online}  & ReIL               & 0.29 (0.061)          & 3.65 & 40.4 (1.783)           & \textbf{0.07 (0.016)}  \\ \cline{3-7} 
                    &                          & ReIL (Only BC)     & 0.322 (0.063)          & 4.5  & 46.17 (3.92)           & 0.081 (0.022)          \\ \cline{3-7} 
                    &                          & ReIL (Pre-trained) & \textbf{0.27 (0.058)}           & 3.58 & \textbf{35.28 (0.918)} & \textbf{0.063 (0.02)}  \\ \cline{3-7} 
                    &                          & HG-DAgger          & 0.285 (0.062)          & 4.58 & 35.64 (0.846)          & 0.14 (0.057)           \\ \cline{2-7} 
                    & \multirow{2}{*}{Offline} & ReIL               & \textbf{0.247 (0.034)}          & 0.27 & 35.73 (1.062)          & \textbf{0.056 (0.009)} \\ \cline{3-7} 
                    &                          & HG-DAgger          & 0.307 (0.055)          & 1.00 & 35.87 (0.46)           & 0.158 (0.022)          \\ \hline
\multirow{4}{*}{TH} & \multirow{2}{*}{Online}  & ReIL               & 0.301 (0.079)          & 2.72 & \textbf{19.98 (1.493)} & 0.086 (0.024)          \\ \cline{3-7} 
                    &                          & ReIL (Only BC)     & 0.317 (0.082)            & 3.06 & 22.55 (3.02)           & 0.085 (0.022)          \\ \cline{2-7} 
                    & \multirow{2}{*}{Offline} & ReIL               & 0.288 (0.058)          & 0.0  & 17.47 (0.61)           & \textbf{0.09 (0.028)}  \\ \cline{3-7} 
                    &                          & HG-DAgger          & 0.277 (0.039)          & 0.0  & 17.73 (0.64)           & 0.128 (0.027)          \\ \hline
\end{tabular}}
\label{table:mobile_robot_result}
\end{center}
\footnotesize{OA = Obstacle avoidance, TH = Target hitting, Mode = Online/Offline, Error = Average action error, Supervised = Average supervised steps per episode, Steps = Average steps per episode, Angular Acc. = Average absolute angular acceleration. Results significantly better than others according to a two-tailed t-test are highlighted in \textbf{bold} for each task and mode of training.}
\end{table}

\section{Results} \label{result}

Results for the simulated cartpole environment are shown in TABLE \ref{table:cartpole_result} and Fig. \ref{fig:cartpole-result}, and the results for the real world visual navigation tasks are shown in TABLE \ref{table:mobile_robot_result} and Fig. \ref{fig:mobile-robot-result}.

We note that for a simple simulated environment such as CartPole-V1, HG-DAgger performs very effectively, despite not being trained on agent-generated data. ReIL was also successful in reaching the goal of 3000 consecutive steps in 20/20 runs and used approximately the same amount of supervised steps as did HG-DAgger. IARL was only 80\% successful in reaching the goal and also required more supervised steps compared to ReIL and HG-DAgger for those successful runs. Finally, ReIL (Only RL), which is essentially L2D, was only successful in 3/4 runs and required a considerably larger amount of supervisor corrections. 

These simple simulation results might lead us to the simple conclusion that methods using BC are more effective than methods using RL; however, in the more complex visual navigation tasks, we find that the performance of methods solely relying on BC deteriorates as the amount of data increases. For ReIL (Only BC), this phenomenon can be seen in Fig. \ref{fig:mobile-robot-result} (a) --- the episode length grows over time as a result of imitating sub-optimal agent-generated data with BC. We note that IWR may also experience this deterioration, as it similarly utilizes BC on both supervisor and agent actions. As for HG-DAgger, Fig. \ref{fig:mobile-robot-result} (b) indicates that the path generated by the agent is increasingly oscillatory as the number of supervisor corrections increases. This behavior is expected, as HG-DAgger does not learn to avoid supervisor interventions, but only learns to make corrections from undesirable states. Users utilizing HG-Dagger may therefore have to balance the amount of BC pre-training, in which the supervisor demonstrates the whole task without the agent, and the amount of IIL supervisor corrections in order to minimize these oscillations.

In contrast, the reinforced method ReIL does not experience deterioration in performance over time (as shown in Fig. \ref{fig:mobile-robot-result} (a) and (b)), as the critic takes into account both agent-generated and supervisor-generated data and assesses the relative quality of each data point according to a reward function. Comparing ReIL and HG-DAgger in the offline training setting (see TABLE \ref{table:mobile_robot_result}), in which they shared the same dataset, we see again the superior performance of ReIL and the viability of training it offline. On a final note, the MimeticSNAIL model utilized in the visual navigation tasks did not require any manual fine tuning, making it simple to train agents on multiple tasks.

\section{Conclusion} \label{conclusion}

In this paper, we describe the design and empirical evaluation of ReIL, a framework for intervention-based imitation learning that combines the advantages of imitation learning and reinforcement learning approaches. Our experimental results on real world navigation tasks indicate that the combination of these two components enhances both the performance and the trainability of the agent in comparison with current work that utilizes only one or the other of these components. Furthermore, we also hypothesize an advantage of ReIL over other similar algorithms such as IARL and find empirical support for that hypothesis in ReIL's performance in the simulated cartpole balancing task. 

In future work, we hope to create an IIL algorithm that utilizes ReIL's value estimator (critic) to predict and propose supervisor interventions \textit{before} the agent makes a mistake. This could simplify training of the actor, as it will no longer need to explore the environment with RL to learn desirable actions that help avoid interventions.

\bibliographystyle{IEEEtran}
\bibliography{References} 

\end{document}